\documentclass{article}

\usepackage{acronym}
\usepackage{amsmath}
\usepackage{microtype}
\usepackage{graphicx}
\usepackage{subfigure}
\usepackage{booktabs} 
\usepackage{algorithm}
\usepackage[noend]{algpseudocode}
\usepackage{hyperref}
\usepackage{cleveref}
\usepackage{makecell}
\usepackage{booktabs}
\usepackage[font=small,skip=0pt]{caption}

\usepackage[accepted]{icml2019}

\icmltitlerunning{CVPR 2020 Continual Learning in Computer Vision Competition}

\begin{document}

\twocolumn[
\icmltitle{CVPR 2020 Continual Learning in Computer Vision Competition: Approaches, Results, Current Challenges and Future Directions}




\begin{icmlauthorlist}
\icmlcorrespondingauthor{Vincenzo Lomonaco}{vincenzo.lomonaco@unibo.it}
\icmlauthor{Vincenzo Lomonaco}{unibo,continualai}
\icmlauthor{Lorenzo Pellegrini}{unibo}
\icmlauthor{Pau Rodriguez}{elementai}
\icmlauthor{Massimo Caccia}{mila,elementai}
\icmlauthor{Qi She}{bytedance,continualai}
\icmlauthor{Yu Chen}{bristol}
\icmlauthor{Quentin Jodelet}{titech,aist}
\icmlauthor{Ruiping Wang}{cas}
\icmlauthor{Zheda Mai}{toronto}
\icmlauthor{David Vazquez}{elementai}
\icmlauthor{German I. Parisi}{hamburg,continualai}
\icmlauthor{Nikhil Churamani}{cambridge}
\icmlauthor{Marc Pickett}{googleai}
\icmlauthor{Issam Laradji}{elementai}
\icmlauthor{Davide Maltoni}{unibo}
\end{icmlauthorlist}

\icmlaffiliation{unibo}{University of Bologna}
\icmlaffiliation{continualai}{ContinualAI Research}
\icmlaffiliation{elementai}{Element AI}
\icmlaffiliation{mila}{MILA}
\icmlaffiliation{hamburg}{University of Hamburg}
\icmlaffiliation{cambridge}{University of Cambridge}
\icmlaffiliation{googleai}{Google AI}
\icmlaffiliation{bytedance}{ByteDance AI Labs}
\icmlaffiliation{bristol}{University of Bristol}
\icmlaffiliation{titech}{Tokyo Institute of Technology}
\icmlaffiliation{aist}{AIST RWBC-OIL}
\icmlaffiliation{cas}{Chinese Academy of Sciences}
\icmlaffiliation{toronto}{University of Toronto}

\icmlkeywords{Continual Learning, Lifelong Learning, Computer Vision, Competition}

\vskip 0.3in
]



\printAffiliationsAndNotice{} 

\begin{abstract}

In the last few years, we have witnessed a renewed and fast-growing interest in continual learning with deep neural networks with the shared objective of making current AI systems more adaptive, efficient and autonomous. However, despite the significant and undoubted progress of the field in addressing the issue of catastrophic forgetting, benchmarking different continual learning approaches is a difficult task by itself. In fact, given the proliferation of different settings, training and evaluation protocols, metrics and nomenclature, it is often tricky to properly characterize a continual learning algorithm, relate it to other solutions and gauge its real-world applicability. The first Continual Learning in Computer Vision challenge held at CVPR in 2020 has been one of the first opportunities to evaluate different continual learning algorithms on a common hardware with a large set of shared evaluation metrics and 3 different settings based on the realistic CORe50 video benchmark. In this paper, we report the main results of the competition, which counted more than 79 teams registered, 11 finalists and 2300\$ in prizes. We also summarize the winning approaches, current challenges and future research directions.

\end{abstract}

\begin{figure}[t]
  \centering
  \includegraphics[width=\columnwidth]{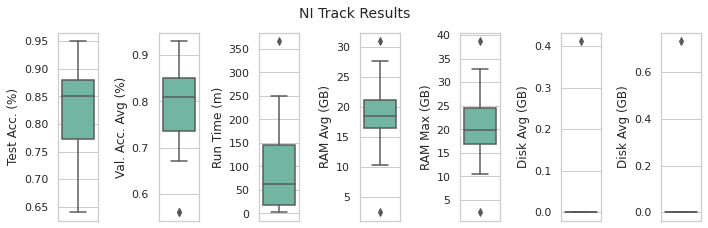}
  \includegraphics[width=\columnwidth]{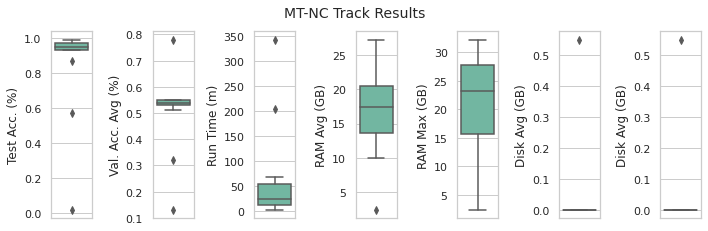}
 \includegraphics[width=\columnwidth]{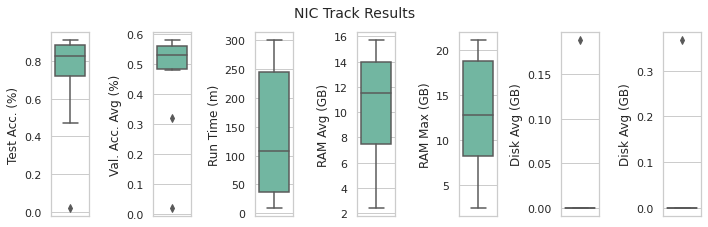}
  \caption{Results distributions for the three tracks (NI, MT-NC and NIC) across the 11 finalists solutions and the main evaluation metrics used for the competition: total test accuracy (\%) at the end of the training, average validation accuracy over time (\%), maximum and average RAM/Disk usage (GB).}
  \label{img:distr_results}
\end{figure}

\section{Introduction}

Continual Learning, the new deep learning embodiment of a long-standing paradigm in machine learning research and AI also known as Incremental or Lifelong Learning, has received a renewed attention from the research community over the last few years \cite{Parisi2019, lomonaco2019a, Lesort2019a}. Continual learning, indeed, appears more and more clearly as the only viable option for sustainable AI agents that can scale efficiently in terms of general intelligence capabilities while adapting to ever-changing environments and unpredictable circumstances over time. Even not considering long-term goals of truly intelligent AI agents, from a pure engineering perspective, continual learning is a very desirable option for any AI technology learning on premises or at the edge on embedded devices without the need of moving private data to remote cloud infrastructures \cite{farquhar2019differentially}.

However, gradient-based architectures, such as neural networks trained with Stochastic Gradient Descent (SGD), notably suffer from catastrophic forgetting or interference \cite{mccloskey1989catastrophic, robins1995catastrophic, french1999catastrophic}, where the network parameters are rapidly overwritten when learning over non-stationary data distributions to model only the most recent. In the last few years, significant progresses have been made to tame the issue. Nevertheless, comparing continual learning algorithms today constitutes a hard task \cite{diaz2018don}. This is mainly due to the proliferation of different settings only covering partial aspects of the continual learning paradigm, with diverse training and evaluation protocols, metrics and datasets used \cite{Lesort2019a,caccia2020online}. Another important question is whether such algorithms, that have mostly been proved on artificial benchmarks such as MNIST \citep{lecun1998gradient} or CIFAR \citep{krizhevsky2009learning}, can scale and generalize to different settings and real-world applications. 

The \emph{1st Continual Learning in Computer Vision Challenge}, organized within the \emph{CLVision} workshop at CVPR 2020, is one of the first attempts to address these questions. In particular, the main objectives of the competition were:

\begin{itemize}
    \item Invite the research community to scale up continual learning approaches to natural images and possibly on video benchmarks.
    \item Invite the community to work on solutions that can generalize over multiple continual learning protocols and settings (e.g. with or without a “task” supervised signal).
    \item Provide the first opportunity for a comprehensive evaluation on a shared hardware platform for a fair comparison.
\end{itemize}

Notable competitions previously organized in this area include: the Pascal 2 EU network of excellence challenge on \emph{“covariate shift”}, organized in 2005 \cite{firstclchallenge, quionero2009dataset}; the \emph{Autonomous Lifelong Machine Learning with Drift} challenge organized at NeurIPS 2018 \cite{escalante2020automl} and the \emph{IROS 2019 Lifelong Robotic Vision} challenge \cite{bae2020iros}. While the first two competitions can be considered as the first continual learning challenges ever organized, they were based on low-dimensional features benchmarks that made it difficult to understand the scalability of the proposed methods to more complex settings with deep learning based techniques. The latest competition, instead, has been one of the first challenges organized within robotic vision realistic settings. However, it lacked a general focus on computer vision applications as well as a comprehensive evaluation on 3 different settings and 4 tracks.

For transparency and reproducibility, we openly release the finalists' dockerized solutions as well as the initial baselines at the following link: \url{https://github.com/vlomonaco/cvpr_clvision_challenge}.

\section{Competition}

\begin{table*}[t]
\caption{11 finalists of the CLVision Competition.}
\label{tab: finalists}
\vskip 0.15in
\begin{center}
\begin{small}
\begin{tabular}{ll}
\toprule
Team Name & Team Members \\
\midrule
\textbf{HaoranZhu} &  Haoran Zhu \\
\textbf{ICT\_VIPL}  & Chen He, Qiyang Wan, Fengyuan Yang, Ruiping Wang, Shiguang Shan, Xilin Chen\\
\textbf{JimiB}     & Giacomo Bonato, Francesco Lakj, Alex Torcinovich, Alessandro Casella\\
\textbf{Jodelet}   & Quentin Jodelet, Vincent Gripon, Tsuyoshi Murata\\
\textbf{Jun2Tong}  & Junyong Tong, Amir Nazemi, Mohammad Javad Shafiee, Paul Fieguth
\\
\textbf{MrGranddy} & Vahit Bugra Yesilkaynak, Firat Oncel, Furkan Ozcelik, Yusuf Huseyin Sahin, Gozde Unal
 \\
\textbf{Noobmaster}  & Zhaoyang Wu, Yilin Shao, Jiaxuan Zhao, and Bingnan Hu \\
\textbf{Sahinyu} & Yusuf H. Sahin, Furkan Ozcelik, Firat Oncel, Vahit Bugra Yesilkaynak, Gozde Unal \\
\textbf{Soony}    &  Soonyong Song, Heechul Bae, Hyonyoung Han, Youngsung Son\\
\textbf{UT\_LG}    &  Zheda Mai, Hyunwoo Kim, Jihwan Jeong, Scott Sanner\\
\textbf{YC14600}    &   Yu Chen, Jian Ma, Hanyuan Wang, Yuhang Ming, Jordan Massiah, Tom Diethe\\
\bottomrule
\end{tabular}
\end{small}
\end{center}
\vskip -0.1in
\end{table*}

The \emph{CLVision competition} was planned as a 2-phase event (pre-selection and finals), with 4 tracks and held online from the \emph{15th of February 2020} to the \emph{14th of June 2020}. 
The \emph{pre-selection} phase, based on the codalab online evaluation framework\footnote{\url{https://codalab.org}}, lasted 78 days and was followed by the finals where a dockerized solution had to be submitted for remote evaluation on a shared hardware. In the following section, the dataset, the different tracks, the evaluation metric used and the main rules of the competition are reported in detail. Finally, the main competition statistics, participants and winners are presented.

\begin{figure}[t]
  \centering
  \includegraphics[width=\columnwidth]{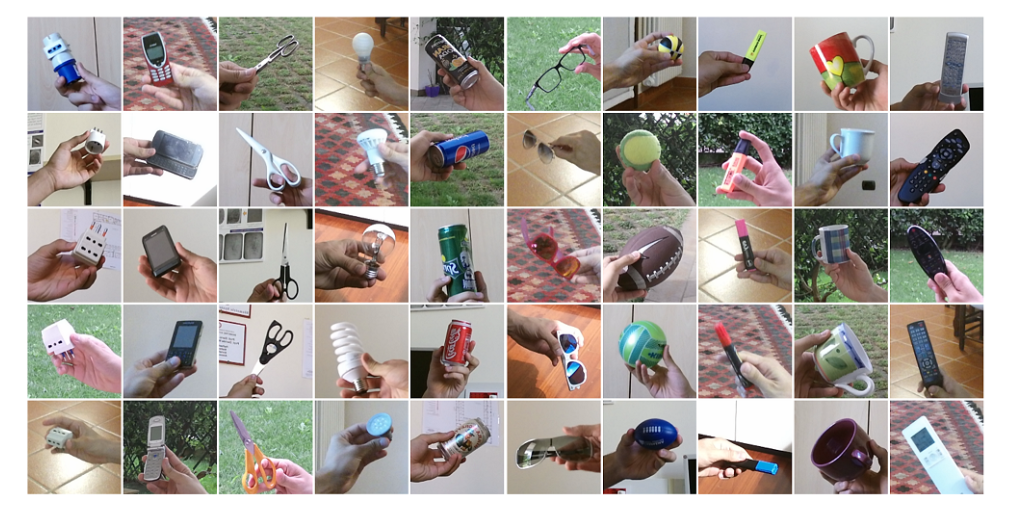}
  \caption{Example images of the 50 objects in \textsl{CORe50}, the main video dataset used in the challenge. Each column denotes one of the 10 categories \cite{Lomonaco2017}.}
  \label{img:core50}
\end{figure}

\subsection{Dataset}

CORe50 \cite{pmlr-v78-lomonaco17a} was  specifically  designed  as  an  object recognition video benchmark for continual learning. It consists of 164,866 128$\times$128 images of 50 domestic objects belonging to 10 categories (see Figure \ref{img:core50}); for each object the dataset includes 11 video sessions ($\sim$300 frames recorded with a Kinect 2 at 20 fps) characterized by relevant variations in terms of lighting, background, pose and occlusions. Classification on CORe50 can be performed at Object level (50 classes) or at Catagory level (10 classes). The former, being a more challenging task, was the configuration chosen for this competition.  The egocentric vision of hand-held objects allows for the emulation of a scenario where a robot has to incrementally learn to recognize objects while manipulating them. Objects are presented to the robot by a human operator who can also provide the labels, thus enabling a supervised classification (such an applicative scenario is well described in \citet{Pasquale2019a, she2019openlorisobject}).

\subsection{Tracks}

Based on the CORe50 dataset, the challenge included four different tracks based on the different settings considered:

\begin{enumerate}
    \item New Instances (NI): In this setting 8 training batches of the same 50 classes are encountered over time. Each training batch is composed of different images collected in different environmental conditions.
    \item Multi-Task New Classes (MT-NC)\footnote{Multi-Task-NC constitutes a simplified variation of the originally proposed New Classes (NC) protocol \cite{pmlr-v78-lomonaco17a} (where the task label is not provided during train and test).}: In this setting the 50 different classes are split into 9 different tasks: 10 classes in the first batch and 5 classes in the other 8. In this case the task label will be provided during training and test.
   \item New Instances and Classes (NIC): this protocol is composed of 391 training batches containing 300 images of a single class. No task label will be provided and each batch may contain images of a class seen before as well as a completely new class.
   \item All together (ALL): All the settings presented above.
\end{enumerate}

Each participant of the challenge could choose in which of the main three tracks (NI, MT-NC, NIC) to compete. Those participants that decided to participate to all the three main tracks were automatically included in the ALL track as well, the most difficult and ambitious track of the competition. 

\subsection{Evaluation Metric}

In the last few years the main evaluation focus in continual learning has always been centered around accuracy-related forgetting metrics. However, as argued by \citet{Diaz-Rodriguez2018}, this may lead to biased conclusion not accounting for the real scalability of such techniques over an increasing number of tasks/batches and more complex settings. For this reason, in the competition each solution was evaluated across a number of metrics:
\begin{enumerate}
    \item \emph{Final accuracy on the test set}\footnote{Accuracy in CORe50 is computed on a fixed test set. Rationale behind this choice is explained in \cite{pmlr-v78-lomonaco17a}}: computed only at the end of the training procedure.
    \item \emph{Average accuracy over time on the validation set}: computed at every batch/task.
    \item \emph{Total training/test time}: total running time from start to end of the main function (in minutes).
    \item \emph{RAM usage}: total memory occupation of the process and its eventual sub-processes. It is computed at every epoch (in MB).
    \item \emph{Disk usage}: only of additional data produced during training (like replay patterns) and additionally stored parameters. It is computed at every epoch (in MB).
\end{enumerate}

The final aggregation metric ($CL_{score}$) is the weighted average of the 1-5 metrics (0.3, 0.1, 0.15, 0.125, 0.125 respectively).

\subsection{Rules and Evaluation Infrastructure}

In order to provide a fair evaluation while not constraining each participants to simplistic solutions due to a limited server-side computational budget, the challenge was based on the following rules: 

\begin{enumerate}
    \item The challenge was based on the \emph{Codalab} platform. For the pre-selection phase, each team was asked to run the experiments \emph{locally} on their machines with the help of a Python repository to easily load the data and generate the submission file (with all the necessary data to execute the submission remotely and verify the adherence to the competition rules if needed). The submission file, once uploaded, was used to compute the $CL_{Score}$ which determined the ranking in each scoreboard (one for each track).
    \item It was possible to optimize the data loader, but not to change the data order or the protocol itself.
    \item The top 11 teams in the scoreboard at the end of the pre-selection phase were selected for the final evaluation.
    \item The final evaluation consisted in a \emph{remote} evaluation of the final submission for each team. This is to make sure the final ranking was computed in the same computational environment for a fair comparison. In this phase, experiments were run remotely for all the teams over a 32 CPU cores, 1 NVIDIA Titan X GPU, 64 GB RAM Linux system. The max running time was capped at 5 hours for each submission/track.
    \item Each team selected for the final evaluation had to submit a single dockerized solution which had to contain the exact same solution submitted for the last codalab evaluation. The initial docker image (provided in the initial challenge repository) could have been customized at will but without exceeding 5 GB.
\end{enumerate}

It is worth noting that only the test accuracy was considered in the ranking of the pre-selection phase of the challenge, since the evaluation was run on participants' local hardware. However, since it was not possible to submit a different solution for the final evaluation, this ensured the competition was not biased on the sole accuracy metric.

The financial budget for the challenge was entirely allocated for the monetary prizes in order to stimulate participation:
\begin{itemize}
    \item 800\$ for the participant with highest average score accross the three tracks (e.g the ALL track).
    \item 500\$ for the participant with highest score on the NI track.
    \item 500\$ for the participant with highest score on the MT-NC track.
    \item 500\$ for the participant with highest score on the NIC track.
\end{itemize}
These prizes were kindly sponsored by \emph{Intel Labs (China)}, while the remote evaluation was performed thanks to the hardware provided by the \emph{University of Bologna}.

\subsection{Participants and Finalists}

The challenge counted the participation of 79 teams worldwide that competed during the pre-selection phase. From those 79 teams only 11 qualified to the finals with a total of 46 people involved and an average team components number of 4. In Table \ref{tab: finalists} the 11 finalist teams and their members are reported.
 
\section{Continual Learning Approaches}

In this section we discuss the baselines made available as well as the continual learning approaches of the winning teams in more details. On the official competition website an extended report for each of the finalist team detailing their approach is also publicly available.\footnote{\url{https://sites.google.com/view/clvision2020/challenge}}

\subsection{Baselines}

In order to better understand the challenge complexity and the competitiveness of the proposed solutions, three main baselines were included for each of the 4 tracks:

\begin{itemize}
    \item \emph{Naive}: This is the basic \emph{finetuning} strategy, where the standard SGD optimization process is continued on the new batches/tasks without any additional regularization constraint, architectural adjustment or memory replay process.
    \item \emph{Rehearsal}: In this baseline the \emph{Naive} approach is augmented with a basic replay process with a growing external memory, where 20 images for each batch are stored.
    \item \emph{AR1* with Latent Replay}: a recently proposed strategy \cite{cloe-preprint} showing competitive results on CORe50 with a shared, non fine-tuned hyper-parametrization across the three main tracks.
\end{itemize}

\subsection{Team ICT\_VIPL}

\textbf{General techniques for all tracks}. To improve their performance the ICT\_VIPL team used: (1) \textit{Heavy Augmentation} with the Python \texttt{imgaug} library\footnote{\url{https://imgaug.readthedocs.io}}; (2) resize the input image to 224$\times$224 to encourage more knowledge transfer from the ImageNet pretrained model; (3) employ an additional exemplar memory for episodic memory replay to alleviate catastrophic forgetting (randomly select $2\sim 3\%$ of the training samples); (4) striking a balance between performance and model capacity by using a moderately deep network ResNet-50. As for efficiency, they leveraged the PyTorch Dataloader module for multi-thread speed-up.

\textbf{Special techniques for individual tracks}. For NI track, there is no special design over the general techniques above and they only tune the best hyper-parameters. For Multi-Task-NC track, they carefully design a pipeline that disentangles representation and classifier learning, which shows very high accuracy and the pipeline is as below ($D_i$ is the set of exemplars for Task $i$ and $|D_i|$ is its size):

\emph{For Task 0}: (1) Train the feature extractor $f(x)$ and the first head $c_0 (z)$ with all training samples; (2) Select N samples randomly and store them in the exemplar memory ($|D_0|=N$).

\emph{For Task $i$} ($i=1,2,\ldots,8$): (1) Train head $c_i(z)$ with all training samples of Task $i$; (2) Drop some samples randomly from the previous memory, keep $|D_j |=\frac{N}{i+1}$ (for all $j<i$); (3) Select $\frac{N}{i+1}$ samples from Task $i$ randomly and store them in the exemplar memory ($|D_i|=\frac{N}{i+1}$); (4) Fine-tune the feature extractor $f(x)$ with all samples in the memory $\cup_{j}D_j (j \leq i)$. (since the feature extractor alone cannot classify images, a temporary head $c(z)$ is used for training); (5) Fine-tune each head $c_j (z)$ with the corresponding samples in the memory $D_j (j\leq i)$.

For NIC track, based on the assumption that the neural network estimates Bayesian a posteriori probilitities~\cite{richard1991neural}, the network outputs are divided by the prior probability for each class inspired by the trick that handles class imbalance~\cite{buda2018systematic}. Such a technique can prevent the classifier from biasing minority class (predict to newly added classes) especially in the first few increments.




\subsection{Team Jodelet}

The proposed solution consists in the concatenation of a pre-trained deep convolutional neural network used as a feature extractor and an online trained logistic regression combined with a small reservoir memory~\cite{chaudhry2019tiny} used for rehearsal.

Since the guiding principle of the proposed solution is to limit as much as possible the computational complexity, the model is trained in an online continual learning setting: each training example is only used once.
In order to further decrease the memory and computational complexity of the solution at the cost of a slight decrease of the accuracy, the pre-trained feature extractor is fixed and is not fine-tuned during the training procedure. As a result, it is not necessary to apply the gradient descent algorithm to the large feature extractor and the produced representation is fixed. Therefore, it is possible to store the feature representation in the reservoir memory instead of the whole input raw image. In addition to the memory gain, this implies that the replay patterns do not have to go through the feature extractor again, effectively decreasing the computational complexity of the proposed solution.

Among the different architectures and training procedures considered for the feature extractor, ResNet-50~\cite{he2016deep} trained by Facebook AI using the Semi-Weakly Supervised Learning procedure~\cite{yalniz2019billion} was selected. This training procedure relies on the use of a teacher model and 940 million public images in addition to the ImageNet dataset~\cite{Russakovsky_2015}. Compared with the reference training procedure in which the feature extractor is solely trained on the ImageNet dataset, this novel training procedure allows for a consequent increase of the accuracy without modifying the architecture: while the difference of Top-1 accuracy between both training procedures for ResNet-50 is about 5.0\% on Imagenet, the difference increases up to 11.1\% on the NIC track of the challenge. Moreover, it should be noted that on the three tracks of the challenge, ResNet-18 feature extractor trained using this new procedure is able to reach an accuracy comparable with the one of the reference ResNet-50 feature extractor trained only on ImageNet, while being considerably smaller and faster.

For reasons of consistency, the same hyperparameters have been used for the three tracks of the challenge and have been selected using a grid search.






\subsection{Team UT\_LG}
\paragraph{Batch-level Experience Replay with Review} 
In most \emph{Experience Replay} based methods, the incoming mini-batch is concatenated with another mini-batch of samples retrieved from the memory buffer. Then, they simply takes an SGD step with the concatenated samples, followed by an update of the memory~\cite{chaudhry2019tiny,caccia2019online}. Team UT\_LG method makes two modifications. Firstly, to reduce the number of retrieval and update steps, they concatenate the memory examples at the batch level instead of at the mini-batch level. Concretely, for every epoch, they draw a batch of data $D_\mathcal{M}$ randomly from memory with size  $replay\_sz$, concatenate it with the current batch and conduct the gradient descent parameters update. Moreover, they add a review step before the final testing, where they draw a batch of size $D_R$ from memory and conduct the gradient update again. To prevent overfitting, the learning rate in the review step is usually lower than the learning rate used when processing incoming batches. The overall training procedure is presented in Algorithm~\ref{alg:algo1}.
\paragraph{Data Preprocessing} (1) Centering-cropping the image with a (100, 100) window to make the target object occupy more pixels in the image. (2) Resizing the cropped image to (224, 224) to ensure no size discrepancy between the input of the pre-trained model and the training images. (3) Pixel-level and spatial-level data augmentation to improve generalization. The details of their implementation can be found in~\cite{mai2020batchlevel}
\begin{algorithm}
\caption{Batch-level Experience Replay with Review}
\label{alg:algo1}
\begin{algorithmic}
\Procedure{BERR}{$\mathcal{D}$, mem\_sz, replay\_sz, review\_sz, lr\_replay, lr\_review}
    \State $\mathcal{M}\leftarrow \{\} * mem\_sz$
    \For{$t\in\{1,\dots,T\}$}
        \For{epochs}
            \If {$t > 1$}
                \State $D_\mathcal{M}\stackrel{\text{replay\_sz}}{\sim}\mathcal{M}$ 
                \State $D_{\text{train}} = D_{\mathcal{M}}\cup D_t$ 
            \Else
                \State $D_{\text{train}} = D_t$
            \EndIf
            \State $\theta \leftarrow$ SGD($D_{\text{train}}, \theta, \text{lr\_replay}$)
        \EndFor
        \State $\mathcal{M}\leftarrow UpdateMemory(D_t, \mathcal{M}, \text{mem\_sz})$ \
    \EndFor
    \State $D_R\stackrel{\text{review\_sz}}{\sim}\mathcal{M}$ 
    \State $\theta\leftarrow \text{SGD}(D_R,\theta,\text{lr\_review})$ 
    \State \textbf{return} $\theta$
\EndProcedure
\end{algorithmic}
\end{algorithm}




\subsection{Team Yc14600}
\acrodef{DRL}{Discriminative Representation Loss}

The use of episodic memories in continual learning is an efficient way to prevent the
phenomenon of \emph{catastrophic forgetting}. In recent studies, several gradient-based approaches have been developed to make more efficient use of compact episodic memories. The essential idea is to use gradients produced by samples from episodic memories to constrain the gradients produced by new samples, \emph{e.g.} by ensuring the inner product of the pair of gradients is non-negative \cite{lopez2017gradient} as follows:
\begin{equation}\label{eq:gem_base}
    \langle g_t,g_k \rangle
    = \left \langle \frac{\partial\mathcal{L}(x_t,\theta)}{\partial \theta}, \frac{\partial\mathcal{L}(x_k,\theta)}{\partial \theta} \right\rangle
    \ge 0, \forall k < t
\end{equation}
where $t$ and $k$ are time indices, $x_t$ denotes a new sample from the current task, and $x_k$ denotes a sample from the episodic memory. Thus, the updates of parameters are forced to preserve the performance on previous tasks as much as possible. \Cref{eq:gem_base} indicates larger cosine similarities between gradients produced by current and previous tasks result in improved generalisation. This in turn indicates that samples that lead to the most diverse gradients provide the most difficulty during learning.

Through empirical studies the team members found that the discrimination ability of representations strongly correlates with the diversity of gradients, and more discriminative representations lead to more consistent gradients. 
They use this insight to introduce an extra objective \acf{DRL} into the optimization objective of classification tasks in continual learning. 
Instead of explicitly refining gradients during training process, \ac{DRL} helps with decreasing gradient diversity by optimizing the representations. As defined in \Cref{eq:drl}, \ac{DRL} consists of two parts: one is for minimizing the similarities of representations between samples from different classes ($\mathcal{L}_{bt}$), the other is for minimizing the similarities of representations between samples from a same class ($\mathcal{L}_{wi}$) for preserving information of representations for future tasks. 
\begin{equation}\label{eq:drl}
     \begin{split}
         & \min_{\Theta}\mathcal{L}_{DR} = \min_{\Theta}(\mathcal{L}_{bt} +  \mathcal{L}_{wi}), \\
         & \mathcal{L}_{bt} = \frac{1}{B_{bt}} \sum_{l=1}^L \sum_{i=1}^B \sum_{j=1,y_j \neq y_i}^B \langle h_{l,i},h_{l,j}\rangle,\\
         & \mathcal{L}_{wi} = \frac{1}{B_{wi}} \sum_{l=1}^L
         \sum_{i=1}^B \sum_{j=1,j \neq i,y_j = y_i}^B \langle h_{l,i},h_{l,j}\rangle.
     \end{split}
 \end{equation}
where $\Theta$ denotes the parameters of the model, $L$ is the number of layers of the model, $B$ is training batch size. $B_{bt}$ and $B_{wi}$ denote the number of pairs of samples in the training batch that are from different classes and the same class, respectively, $h_{l,i}$ is the output of layer $l$ by input $x_i$ and $y_i$ is the label of $x_i$. Please refer to \cite{chen2020bypassing} for more details.

\begin{figure}[th]
  \centering
  \includegraphics[width=\columnwidth]{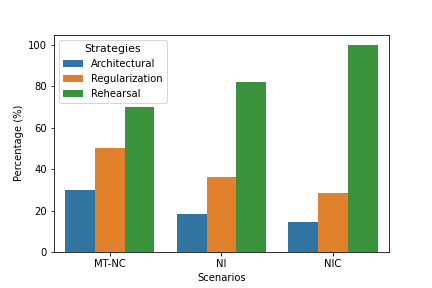}
  \caption{Percentage (\%) of finalists solutions for each track employing an \emph{architectural}, \emph{regularization} or \emph{rehearsal} strategy. Percentages do not sum to 100\% since many approached used hybrid strategies. Better viewed in colors.}
  \label{img:strat_perc}
\end{figure}

\section{Competition Results}

In this section we detail the main results of the competition for each of the main three tracks (NI, MT-NC \& NIC) as well as the averaged track ALL, which determined the overall winner of the challenge. For each track the teams are ranked as follows: i) each metric is normalized across between 0 and 1; ii) the $CL_{score}$ is computed as a weighted average; ii) results are ordered in descending order.

In the next sections we report the results with their absolute values to better grasp the quality of the solutions proposed and their portability in different applicative contexts.

\begin{table*}[th]
\caption{NI track results for the 11 finalists of the competition and the three baselines.}
\label{tab:ni-results}
\vskip 0.15in
\begin{center}
\begin{footnotesize}
\begin{sc}
\setlength{\tabcolsep}{6pt}
\begin{tabular}{lcccccccc}
\toprule
team name	& \makecell{test acc \\(\%)} & \makecell{val acc$_{avg}$ \\(\%)} & \makecell{run$_{time}$ \\(m)} & \makecell{ram$_{avg}$ \\ (mb)} & \makecell{ram$_{max}$ \\(mb)} & \makecell{disk$_{avg}$ \\ (mb)} & \makecell{disk$_{max}$ \\(mb)} & $CL_{score}$ \\
\midrule
UT\_LG	&	0.91 & 0.90 &	63.78 &	11429.83	& 11643.63	&0	&0	&0.692\\
Yc14600	&0.88	&0.85	&22.58	&17336.38	&18446.90	&0	&0	&0.648\\
ICT\_VIPL	&0.95	&0.93	&113.70	&2459.42	&2460.16	&421.875	&750	&0.629\\
Jodelet	&0.84	&0.85	&3.11	&18805.60	&18829.96	&0	&0	&0.612\\
Soony	&0.85	&0.81	&25.57	&16662.73	&17000.10	&0	&0	&0.602\\
JimiB	&0.91	&0.89	&248.82	&19110.84	&25767.74	&0	&0	&0.573\\
Jun2tong &0.84	&0.76	&62.48	&20968.43	&23252.39	&0	&0	&0.550\\
Sahinyu	& 0.88	&0.81	&156.64	&26229.77	&32176.76	&0	&0	&0.538\\
Ar1	& 0.75	&0.73	&17.18	&10550.61	&10838.79	&0	&0	&0.520\\
Noobmaster	& 0.85	&0.75	&74.54	&31750.19	&39627.31	&0	&0	&0.504\\
MrGranddy	& 0.88	&0.84	&249.28	&28384.06	&33636.52	&0	&0	&0.501\\
Naïve	& 0.66	&0.56	&2.61	&18809.50	&18830.11	&0	&0	&0.349\\
Rehearsal	& 0.64	&0.56	&3.79	&21685.03	&21704.76	&0	&0	&0.326\\
HaoranZhu	& 0.70	&0.67	&366.22	&21646.78	&21688.30	&0	&0	&0.263\\
\midrule
avg & 0.82	&0.78	&100.74	&18987.80	&21135.96	&30.13	&53.57	&0.52 \\
\bottomrule
\end{tabular}
\end{sc}
\end{footnotesize}
\end{center}
\vskip -0.1in
\end{table*}

\begin{table*}[t]
\caption{NC track results for the 11 finalists of the competition and the three baselines. Teams not appearing in the table did not compete in this track.}
\label{tab:nc-results}
\vskip 0.15in
\begin{center}
\begin{small}
\begin{sc}
\begin{tabular}{lcccccccc}
\toprule
team name	& \makecell{test acc \\(\%)} & \makecell{val acc$_{avg}$ \\(\%)} & \makecell{run$_{time}$ \\(m)} & \makecell{ram$_{avg}$ \\ (mb)} & \makecell{ram$_{max}$ \\(mb)} & \makecell{disk$_{avg}$ \\ (mb)} & \makecell{disk$_{max}$ \\(mb)} & $CL_{score}$ \\
\midrule

Ar1	&0.93	&0.53	&16.02	&10263.19	&14971.72	&0	&0	&0.693\\
UT\_LG	&0.95	&0.55	&19.02	&13793.31	&16095.20	&0	&0	&0.691\\
Yc14600	&0.97	&0.54	&11.81	&15870.62	&19403.57	&0	&0	&0.686\\
Soony	&0.97	&0.55	&55.02	&14005.91	&16049.12	&0	&0	&0.679\\
Jodelet	&0.97	&0.55	&2.55	&17893.58	&23728.84	&0	&0	&0.679\\
Jun2tong	&0.96	&0.55	&28.80	&18488.68	&19588.57	&0	&0	&0.671\\
ICT\_VIPL	&0.99	&0.55	&25.20	&2432.56	&2432.84	&562.5	&562.5	&0.630\\
Rehearsal	&0.87	&0.51	&4.49	&20446.93	&28329.14	&0	&0	&0.626\\
JimiB	&0.95	&0.78	    &204.56	&21002.95	&24528.27	&0	&0	&0.607\\
MrGranddy	&0.94	&0.54	&46.52	&27904.55	&32921.94	&0	&0	&0.604\\
Noobmaster	&0.95	&0.53	&68.07	&27899.86	&32910.23	&0	&0	&0.597\\
HaoranZhu	&0.57	&0.32	&343.50	&21223.30	&28366.48	&0	&0	&0.351\\
Naïve	&0.02	&0.13	&3.41	&17897.38	&23726.40	&0	&0	&0.318\\
\midrule
avg & 0.85	&0.51	&63.77	&17624.83	&21773.26	&43.27	&43.27	&0.60 \\

\bottomrule
\end{tabular}
\end{sc}
\end{small}
\end{center}
\vskip -0.1in
\end{table*}

\begin{table*}[th]
\caption{NIC track results for the 11 finalists of the competition and the three baselines. Teams not appearing in the table did not compete in this track.}
\label{tab:nic-results}
\vskip 0.15in
\begin{center}
\begin{small}
\begin{sc}
\begin{tabular}{lcccccccc}
\toprule
team name	& \makecell{test acc \\(\%)} & \makecell{val acc$_{avg}$ \\(\%)} & \makecell{run$_{time}$ \\(m)} & \makecell{ram$_{avg}$ \\ (mb)} & \makecell{ram$_{max}$ \\(mb)} & \makecell{disk$_{avg}$ \\ (mb)} & \makecell{disk$_{max}$ \\(mb)} & $CL_{score}$ \\
\midrule

UT\_LG	&0.91	&0.58	&123.22	&6706.61	&7135.77	&0	&0	&0.706\\
Jodelet	&0.83	&0.54	&14.12	&10576.67	&11949.16	&0	&0	&0.694\\
Ar1	&0.71	&0.48	&28.19	&3307.62	&4467.64	&0	&0	&0.693\\
ICT\_VIPL	&0.90	&0.56	&91.29	&2485.95	&2486.03	&192.187	&375	&0.625\\
Yc14600	&0.89	&0.57	&160.24	&16069.91	&21550.97	&0	&0	&0.586\\
Rehearsal	&0.74	&0.50	&60.32	&15038.34	&19488.43	&0	&0	&0.585\\
Soony	&0.82	&0.52	&280.39	&12933.28	&14241.57	&0	&0	&0.533\\
JimiB	&0.87	&0.56	&272.98	&13873.04	&21000.51	&0	&0	&0.533\\
Noobmaster	&0.47	&0.32	&300.15	&14492.13	&18262.32	&0	&0	&0.346\\
Naïve	&0.02	&0.02	&9.45	&10583.50	&11917.55	&0	&0	&0.331\\
\midrule
avg & 0.72	&0.47	&134.03	&10606.70	&13249.99	&19.22	&37.50	&0.56 \\
\bottomrule
\end{tabular}
\end{sc}
\end{small}
\end{center}
\vskip -0.1in
\end{table*}

\begin{table*}[th]
\caption{ALL track results for the 11 finalists of the competition and the three baselines. Teams not appearing in the table did not compete in this track.}
\label{tab:all-results}
\vskip 0.15in
\begin{center}
\begin{small}
\begin{sc}
\begin{tabular}{lcccccccc}
\toprule
team name	& \makecell{test acc \\(\%)} & \makecell{val acc$_{avg}$ \\(\%)} & \makecell{run$_{time}$ \\(m)} & \makecell{ram$_{avg}$ \\ (mb)} & \makecell{ram$_{max}$ \\(mb)} & \makecell{disk$_{avg}$ \\ (mb)} & \makecell{disk$_{max}$ \\(mb)} & $CL_{score}$ \\
\midrule

UT\_LG	&	0.92	&0.68	&68.67	&10643.25	&11624.87	&0	&0	&0.694\\
Jodelet	&	0.88	&0.64	&6.59	&15758.62	&18169.32	&0	&0	&0.680\\
Ar1	&	0.80	&0.58	&20.46	&8040.47	&10092.72	&0	&0	&0.663\\
Yc14600	&	0.91	&0.65	&64.88	&16425.64	&19800.48	&0	&0	&0.653\\
ICT\_VIPL	&	0.95	&0.68	&76.73	&2459.31	&2459.68	&392.187	&562.5	&0.617\\
Soony	&	0.88	&0.63	&120.33	&14533.97	&15763.60	&0	&0	&0.612\\
Rehearsal	&	0.75	&0.52	&22.87	&19056.77	&23174.11	&0	&0	&0.570\\
JimiB	&	0.91	&0.74	&242.12	&17995.61	&23765.51	&0	&0	&0.542\\
Noobmaster	&	0.76	&0.53	&147.59	&24714.06	&30266.62	&0	&0	&0.464\\
Naïve	&	0.23	&0.24	&5.16	&15763.46	&18158.02	&0	&0	&0.327\\
\midrule
avg & 0.80	&0.59	&77.54	&14539.12	&17327.49	&39.22	&56.25	&0.58\\
\bottomrule
\end{tabular}
\end{sc}
\end{small}
\end{center}
\vskip -0.1in
\end{table*}

\subsection{New Instances (NI) Track}

In Tab. \ref{tab:ni-results} the main results for the \emph{New Instances} (NI) track are reported. In Tab. \ref{tab:ni-details}, additional details (not taken into account for the evaluation) for each solution are shown. In this track, the \emph{UT\_LG} obtained the best $CL_{Score}$ with a small gap w.r.t. its competitors. The test accuracy tops 91\% for the winning team, showing competitive performance also in real-world non-stationary applications. It is worth noting that the top-4 solutions all employed a rehearsal-based technique, only in one case supported by a regularization counterpart.

\subsection{Multi-Task NC (MT-NC) Track}

For the MT-NC track, results are reported in Tab. \ref{tab:nc-results} and additional details in Tab. \ref{tab:nc-details} of the Appendix. In this scenario, arguably the easiest since it provided an additional supervised signal (the Task label) the AR1 baseline resulted as the best scoring solution. In fact, while achieving lower accuracy results than the other top-7 solutions, it offered a more efficient algorithmic proposal in terms of both memory and computation (even without a careful hyper-parametrization). It is also interesting to note that, in this scenario, it is possible to achieve impressive accuracy performance ($\sim$99\%) within reasonable computation and memory constraints as shown by the ICT\_VIPL team, the only solution who opted for a disk-based exemplars memorization.

\subsection{New Instances (NIC) Track}

The NIC track results are reported in Tab. \ref{tab:nic-results}. Additional details of each solution are also made available in Tab. \ref{tab:nic-details}. Only 7 over 11 finalist teams submitted a solution for this track. In this case, it is possible to observe generally lower accuracy results and an increase in the running times across the 391 batches.

\subsection{All (ALL) Track}

Finally in Tab. \ref{tab:all-results} the results averaged across tracks are reported for the ALL scoreboard. Also in this case the competing teams were 7 over a total of 11 with \emph{UT\_LG} as the winning team. With an average testing accuracy of $\sim$92\%, a average memory consumption of $\sim$10 GB and a running time of $\sim$68 minutes, its relatively simple solution suggests continual learning for practical object recognition applications to be feasible in the real-world, even with a large number of small non-i.i.d. bathes. 

\subsection{Discussion}

Given the main competition results and the additional solutions details reported in Appendix \ref{sec:details}, we can formulate a number of observations to better understand current issues, consolidated approaches and possible future directions for competitive continual learning algorithms tested on real-world computer vision applications. 

In particular, we note:

\begin{itemize}
    \item \emph{Different difficulty for different scenarios}: averaging the 11 finalists test accuracy results we can easily deduce that the MT-NC track or scenario was easier than the NI one ($\sim$85\% vs $\sim$82\%), while the NIC track was the most difficult with a average accuracy of $\sim$72\%. This is not totally surprising, considering that the MT-NC setting allows access to the additional task labels and the NI scenario does not include dramatic distributional shifts, while the NIC one includes a substantially larger number of smaller training batches. Moreover, a number of researchers already pointed out how different training/testing regimes impacts forgetting and the continual learning process \citep{mirzadeh2020understanding, maltoni2019continuous, hayes2018new}.
    \item \emph{100\% of the teams used a pre-trained model}: All the solutions, for all the tracks started from a pre-trained model on ImageNet. While starting from a pre-trained model is notably becoming a standard for real-world computer vision applications, we find it interesting to point out such a pervasive use in the challenge. While this does not mean pre-trained model should be used for every continual learning algorithm in general, it strongly suggests that for solving real-world computer vision application today, pre-training is mostly needed.
    \item \emph{$\sim$90\% of the teams used a rehearsal strategy}: rehearsal constitutes today one of the easiest and effective solution to continual learning where previous works \citep{hayes2019memory} have shown that even a very small percentage of previously encountered training data can have huge impacts on the final accuracy performance. Hence, it is not surprising that a large number of teams opted to use it for maximizing the $CL_{score}$, which only slightly penalized its usage.
    \item \emph{$\sim$45\% of the teams used a regularization approach}: regularization strategies have been extensively used in the competition. It worth noting though, that only 1 team used it alone and not in conjunction with a plain rehearsal or architectural approaches. 
    \item \emph{only $\sim$27\% of the teams used an architectural approach}: less then one third of the participants did use an architectural approach but only on conjunction with a rehearsal or regularization one. This evidence reinforces the hypothesis that architectural-only approaches are difficult to scale efficiently over a large number of tasks or batches \citep{rusu2016progressive}.
    \item \emph{Increasing replay usage with track complexity}: as shown in Fig. \ref{img:strat_perc}, it is worth noting that as the track complexity increased, the proposed solutions tended to include more replay mechanisms. For example, for the NIC track, all the approaches included rehearsal, often used in conjunction with a regularization or architectural approach. 
    \item \emph{High memory replay size}: it is interesting to note that many CL solutions employing rehearsal have chosen to use a \emph{growing} memory replay buffer rather than a fixed one with an average maximum memory size (across teams and tracks) of $\sim$26k patterns. This is a very large number considering that is about $\sim$21\% of the total CORe50 training set images.
    \item \emph{Different hyper-parameters selection}: An important note to make is about the hyperparameters selection and its implication to algorithms \emph{generalization} and \emph{robustness}. Almost all participants' solutions involved a carefully fine-tuned hyper-parameters selection which was different based on the continual scenario tackled. This somehow highlights the weakness of state-of-the-art algorithms and their inability to truly generalize to novel situations never encountered before. A notably exception is the \emph{AR1} baseline, which performed reasonably well in all the tracks with a shared hyperparametrization.

\end{itemize}

\section{Conclusions and Future Improvements}

The \emph{1st Continual Learning for Computer Vision Challenge} held at CVPR2020 has been one of the first large-scale continual learning competition ever organized with a raised benchmark complexity and targeting real-word applications in computer vision. This challenge allowed every continual learning algorithm to be fairly evaluated with shared and unifying criteria and pushing the CL community to work on more realistic benchmarks than the more common MNIST or CIFAR.

After a carefully investigation and analysis of the competition results we can conclude that continual learning algorithms are mostly ready to face real-world settings involving high-dimensional video streams. This is mostly thanks to hybrid approaches often combined with plain replay mechanisms. However, it remains unclear if such techniques can scale over longer data sequences and without such an extensive use of replay. 

Despite the significant participation and success of the 1st edition of the challenge, a number of possible improvements and suggestions for future continual learning competitions can be formulated:

\begin{itemize}
    \item \emph{Discourage over-engineered solutions}: one of the main goal of the competition was to evaluate the applicability of current continual learning algorithms on real-world computer vision problems. However, given the substantial freedom given through the competition rules to achieve this goal, we have noticed a number of over-engineered solutions aimed at improving the $CL_{score}$ but not really significant in terms of novelty of scientific interest. This in turns forced every other participants to focus on over-engineering rather than on the core continual learning issues. For example, data loading or compression algorithms may be useful to decrease memory and compute overheads but may be applicable to most of the solutions proposed, making them less interesting and out of the scope of competition. For this reason, we believe that finding a good trade-off between realism  and scientific interest of the competition will be fundamental for future challenges in this area. We suggest for example to block the possibility to optimize the data loading algorithms and to count the number of replay patterns rather than their bytes overhead.
    \item \emph{Automatize evaluation}: in the current settings of the challenge the evaluation was client-side (on the participants machines) for the pre-selection phase and on a server-side shared hardware for the finals. To ensure the fairness of the results and the competition rules adherence, the code that generated each submission had to be included as well. However, an always-available remote docker evaluation similar to the one proposed for the AnimalAI Olympics \citep{crosby2019animal}, would allow a single phase competition with an always coherent and updated scoreboard, stimulating in turns teams participation and retention over the competition period. This would also alleviate some burdens at the organization levels, reducing the amount of manual interventions.
    \item \emph{Add scalability metrics}: An interesting idea to tame the challenge complexity while still providing a good venue for assessing continual learning algorithms advancement, would be to include other than the already proposed metrics, a number of derivative ones taking into account their trend over time rather than their absolute value. This would help to better understand their scalability on more complex problems and longer tasks/batches sequences and incentivize efficient solutions with constant memory/computation overheads.
    \item \emph{Encourage the focus on original learning strategies}: Another important possible improvement of the competition would be setting up a number of incentives and disincentives to explore interesting research directions in continual learning. For example, the usage of pre-trained models has been extensively used for the competition by all the participants. However it would have been also interesting to see proposals not taking advantage of it as well. In the next competition edition we plan to discourage the use of pre-trained models, different hyperparameters for each setting track and increase the memory usage weight associated to the $CL_{score}$.
\end{itemize}


\section*{Acknowledgements}

We would like to thank all the \emph{Continual Learning in Computer Vision} workshop organizers, challenge chairs and participants for making this competition possible. We also like to acknowledge our sponsors \emph{ContinualAI}, \emph{Element AI}, \emph{Nvidia} and \emph{Intel Labs} for their support in the organization of the workshop at CVPR 2020.


\bibliography{library}
\bibliographystyle{icml2019}


\onecolumn
\appendix
\section{Additional Details}
\label{sec:details}

In this appendix, additional details for each team and track are reported (see Tab. \ref{tab:ni-details}, Tab. \ref{tab:nc-details} and Tab. \ref{tab:nc-details}). In particular we report: i) the model type; ii) if the model was pre-trained; iii) the type of strategy used; iv) the number of eventual replay examples; v) the number of training epochs per batch; vi) the mini-batch size used. 

\begin{table}[!th]
\setlength{\tabcolsep}{1pt}
\caption{Approaches and baselines details for the NI track.}
\label{tab:ni-details}
\vskip 0.1in
\renewcommand{\arraystretch}{1}
\begin{center}
\begin{small}
\begin{tabular}{lcccccc}
\toprule
Team & Model	& Pre-trained & Strategy	&Replay Examples$_{max}$	& Epochs	& Mini-batch size\\
\midrule
UT\_LG & DenseNet-161 & yes	& rehearsal & 80000	&2	&32\\
Yc14600 & ResNeSt50 &	yes &	regularization \& rehearsal	&12000&	1	&16\\
ICT\_VIPL & WideResNet-50	&yes &	rehearsal&	4000&	2&	80\\
Jodelet & ResNet-50	& yes	& rehearsal &	6400&	1&	32\\
Soony & ResNext101/50 \& DenseNet161&	yes &	architectural \& rehearsal	&119894	&1	&900\\
JimiB & resnext101	&yes &	regularization \& rehearsal &	11989&	8&	32\\
Jun2tong & ResNet-50&	yes	& regularization \& rehearsal &12000&	5&	32\\
Sahinyu & Efficientnet-b7&	yes&  rehearsal &	8000&	2&	27\\
Ar1 & mobilenetV1&	yes&	architectural &	1500	&4	&128\\
Noobmaster & resnet-101&	yes&	rehearsal &	24000	&5	&32\\
MrGranddy & EfficientNet-B7&	yes&	regularization \& architectural	&0	&1	&32\\
Naïve & mobilenetV1 &	yes &	n.a.	& 0	& 4	& 128\\
Rehearsal & mobilenetV1&	yes&	rehearsal &	160	&4	&128\\
HaoranZhu & ResNet-50&	yes&	regularization &	0&	10&	32\\

\bottomrule
\end{tabular}
\end{small}
\end{center}
\vskip -0.1in
\end{table}

\begin{table}[!h]
\setlength{\tabcolsep}{1pt}
\caption{Approaches and baselines details the MT-NC track.}
\label{tab:nc-details}
\vskip 0.1in
\renewcommand{\arraystretch}{1}
\begin{center}
\begin{small}
\begin{tabular}{lcccccc}
\toprule
Team & Model	& Pre-trained & Strategy	&Replay Examples$_{max}$	& Epochs	& Mini-batch size\\
\midrule
Ar1	& mobilenetV1 & yes	& architectural \& rehearsal &	1500	&4	&128\\
UT\_LG	&	DenseNet-161&	yes	& architectural	&0	&1	&32\\
Yc14600	&	ResNeSt50&	yes& regularization \& rehearsal &	4500&	1&	16\\
Soony	&	ResNext101/50 \& DenseNet161&	yes& architectural \& rehearsal&	119890&	3&	100\\
Jodelet	&	ResNet-50&	yes&	rehearsal	&6400	&1	&32\\
jun2tong	&	ResNet-50&	yes&	regularization \& rehearsal	&45000	&1	&32\\
ICT\_VIPL	&	ResNeXt-50&	yes&	rehearsal&	3000&	1&	32\\
Rehearsal	&	mobilenetV1&	yes&	rehearsal&	180&	4&	128\\
JimiB	&	resnext101&	yes&	regularization \& rehearsal &	11989&	8&	32\\
MrGranddy	&	EfficientNet-B7&	yes&	regularization \& architectural &	0	&1	&32\\
Noobmaster	&	resnet-101&	yes&	rehearsal &	18000	&5	&32\\
HaoranZhu	&	ResNet-50&	yes&	regularization	&0	&10	&32\\
Naïve	&	mobilenetV1&	yes&	n.a.&	0&	4&	128\\

\bottomrule
\end{tabular}
\end{small}
\end{center}
\vskip -0.1in
\end{table}

\begin{table}[!h]
\setlength{\tabcolsep}{1pt}
\caption{Approaches and baselines details for the NIC track.}
\label{tab:nic-details}
\vskip 0.1in
\renewcommand{\arraystretch}{1}
\begin{center}
\begin{small}
\begin{tabular}{lcccccc}
\toprule
Team & Model	& Pre-trained & Strategy	&Replay Examples$_{max}$	& Epochs	& Mini-batch size\\
\midrule

UT\_LG	&	DenseNet-161&	yes&	rehearsal	&78200&	1&	32\\
Jodelet	&	ResNet-50&	yes&	rehearsal &	6400&	1&	32\\
Ar1	&	mobilenetV1&	yes&	architectural \& rehearsal&	1500&	4&	128\\
ICT\_VIPL	&	ResNet50&	yes&	rehearsal	&2000	&1	&64\\
Yc14600	&	ResNeSt50&	yes&	regularization \& rehearsal	&19550	&1	&32\\
Rehearsal	&	mobilenetV1&	yes&	rehearsal	&7820	&4	&128\\
Soony	&	ResNext101/50 \& DenseNet161 &	yes&	architectural \& rehearsal &	119890&	1&	900\\
JimiB &	resnext101&	yes&	regularization \& rehearsal	&11989	&6	&32\\
Noobmaster &	resnet-101&	yes&	rehearsal	&23460	&5	&32\\
Naïve & mobilenetV1&	yes&	n.a.	&0	&4 &	128\\
\bottomrule
\end{tabular}
\end{small}
\end{center}
\vskip -0.1in
\end{table}

\end{document}